%% file: main.tex
\documentclass{article}

\usepackage[preprint, nonatbib]{neurips_2023}

\usepackage[utf8]{inputenc} 
\usepackage[T1]{fontenc}    
\usepackage{url}            
\usepackage{booktabs}       
\usepackage{amsfonts}       
\usepackage{nicefrac}       
\usepackage{microtype}      
\usepackage{xcolor}         
\usepackage{makecell} 
\usepackage{adjustbox}
\usepackage{pifont}
\usepackage{bbding}
\usepackage{tabulary,multirow,xspace}
\usepackage{fixmath,mathtools,nicefrac,mmstyle}
\usepackage{subcaption}
\usepackage{caption}
\usepackage{float}
\usepackage{wrapfig} 
\usepackage[misc]{ifsym} 
\usepackage{colortbl}
\definecolor{mygray1}{gray}{.95}
\definecolor{mygray2}{gray}{.9}
\definecolor{mygray3}{gray}{.95}
\usepackage{pifont}
\usepackage{newfloat}
\usepackage{listings}

\newlength\savewidth
\newcolumntype{x}[1]{>{\centering\arraybackslash}p{#1pt}}

\newcommand{\app}{\raise.17ex\hbox{$\scriptstyle\sim$}}

\definecolor{linkcolor}{RGB}{255,0,0}
\definecolor{urlcolor}{RGB}{255,105,180}
\definecolor{citecolor}{RGB}{66,168,235}
\usepackage[pagebackref=true,breaklinks=true,letterpaper=true,colorlinks,bookmarks=false]{hyperref}
\hypersetup{colorlinks=true,linkcolor=linkcolor,urlcolor=urlcolor,citecolor=blue}

\newcommand \footnoteONLYtext[1]
{
	\let \mybackup \thefootnote
	\let \thefootnote \relax
	\footnotetext{#1}
	\let \thefootnote \mybackup
	\let \mybackup \imareallyundefinedcommand
}

\title{SU-SAM: A Simple Unified Framework for Adapting Segment Anything Model in Underperformed Scenes}

\author{Yiran Song$^{1}$\footnotemark[1], 
        Qianyu Zhou$^{1}$\thanks{The first two authors contribute equally to this work.}, 
        Xuequan Lu$^{2}$, 
        \textbf{Zhiwen Shao$^{1}$},
        \textbf{Lizhuang Ma$^{1}$} \\
  \textbf{{$^{1}$Shanghai Jiao Tong University}
  {$^{2}$La Trobe University}}\\
  Code: \url{https://github.com/zongzi13545329/SimAda}
}

\begin{document}

\maketitle

\input{latex/0_abstract}
\input{latex/1_introduction}
\input{latex/2_related_work}
\input{latex/3_method}
\input{latex/4_experiment}
\input{latex/5_conclusion}

\clearpage
{\small
  \bibliographystyle{ieee_fullname}
  \bibliography{refbib}
}
\clearpage
\appendix
\input{latex/6_supple}

\end{document}

%% file: latex/0_abstract.tex
\begin{abstract}
Segment anything model (SAM) has demonstrated excellent generalization capabilities in common vision scenarios, yet falling short of the ability to understand specialized data. Recently, several methods have combined parameter-efficient techniques with task-specific designs to fine-tune SAM on particular tasks. However, these methods heavily rely on handcraft, complicated, and task-specific designs, and pre/post-processing to achieve acceptable performances on downstream tasks. As a result, this severely restricts generalizability to other downstream tasks. To address this issue, we present a simple and unified framework, namely SU-SAM, that can easily and efficiently fine-tune the SAM model with parameter-efficient techniques while maintaining excellent generalizability toward various downstream tasks. Our SU-SAM does not require any task-specific designs and aims to improve the adaptability of SAM-like models toward underperformed scenes significantly. Concretely, we abstract the parameter-efficient modules of different methods into basic design elements in our framework. Besides, we propose four variants of SU-SAM, \emph{i.e.,} series, parallel, mixed, and LoRA structures.  We conduct comprehensive experiments on nine datasets and six downstream tasks to verify the effectiveness of SU-SAM, including medical image segmentation, camouflage object detection, salient object segmentation, surface defect segmentation, complex object shapes, and shadow masking. Our experimental results demonstrate that SU-SAM achieves competitive or superior accuracy compared to state-of-the-art methods. Furthermore, we provide in-depth analyses highlighting the effectiveness of different parameter-efficient designs within SU-SAM. In addition, we propose a generalized model and benchmark, showcasing SU-SAM's generalizability across all diverse datasets simultaneously. Our presented SU-SAM offers a flexible and efficient solution for adapting SAM-like models to various downstream tasks.
\end{abstract}

%% file: latex/1_introduction.tex
  \begin{figure*}
    \includegraphics[width=\textwidth]{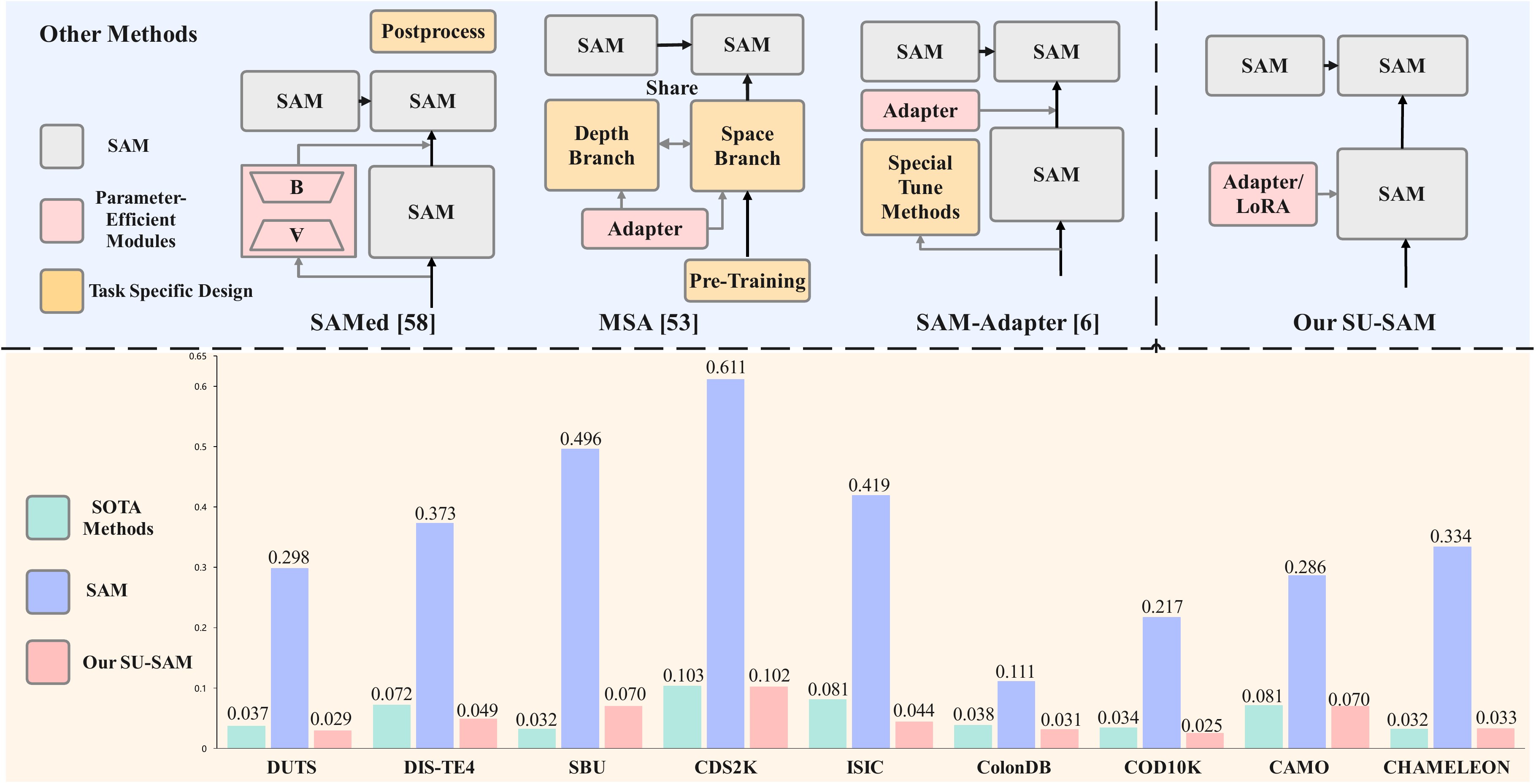}
    \vspace{-4mm}
    \captionof{figure}{\textbf{Top}: contrast between prior SAM-based Paramter-efficient fine-tuning (PEFT) methods~\cite{wu2023medical, zhang2023customized,chen2023sam} and our SU-SAM. Previous works combined task-specific designs to pursue higher accuracy. Unlike them, our SU-SAM removes all task-related operations. \textbf{Bottom}: extensive experiments on nine datasets covering six tasks demonstrate that the proposed SU-SAM can significantly improve the performance of SAM. The evaluation metric is MAE, and a lower value indicates better performance.  
    }
    \label{teaser}
  \end{figure*}

\section{Introduction}
\label{sec:intro}
Recently, there has been a growing interest in foundation models due to their extensive pre-training on large-scale datasets and superior generalization capabilities across various downstream tasks. In the field of computer vision and multimedia~\cite{he2016deep,chen2021exploring,devlin2018bert,brown2020language,raffel2020exploring,CLIP,ramesh2022hierarchical}, 
Segment Anything Model (SAM)~\cite{SAM} has recently gained wide attention. 
As a large-scale vision segmentation model trained on the SA-1B dataset, it utilizes a single user interface as a prompt for segmentation without additional training and achieves good accuracy on general datasets. SAM has shown promising potential in many real-world applications, however, due to the large parameter scale, it is time-consuming and cost-expensive to retrain or fine-tune the whole SAM model for each downstream task or scenario.

Parameter-Efficient Fine-tuning (PEFT), an emerging technique that allows model adaptation without extensively modifying the network and parameters, has been recently studied to improve the adaptability toward various downstream tasks. As illustrated in Figure~\ref{teaser}, several lightweight alternatives~\cite{wu2023medical,chen2023sam,zhang2023customized} have recently been proposed to improve the SAM's performance in specific tasks, which combines parameter-efficient modules with task-specific designs. 
Specifically, SAMed~\cite{zhang2023customized} post-processes the data and incorporates a LoRA module for fine-tuning. 
MSA~\cite{wu2023medical} utilizes a dual-branch structure (Depth Branch and Space Branch) for medical data segmentation, along with additional pre-training steps. 
SAM-Adapter~\cite{chen2023sam} integrates Special Tune Methods with Adapter design, enhancing segmentation accuracy by introducing edge extraction methods. Despite their gratifying progress on specific tasks, these methods heavily rely on many handcraft and complicated designs and pre/post-processing, \emph{e.g.,} carefully-designed data augmentations, additional auxiliary input information, and post-processing of outputs, to achieve desirable performance on the specific downstream task. Such methods would inevitably exhibit performance degradation in other downstream tasks due to the lack of consideration of unified task designs,  compromising the model's ability when generalizing to other downstream tasks.

\begin{figure}[t!]
\centering
\includegraphics[width=0.8\columnwidth]{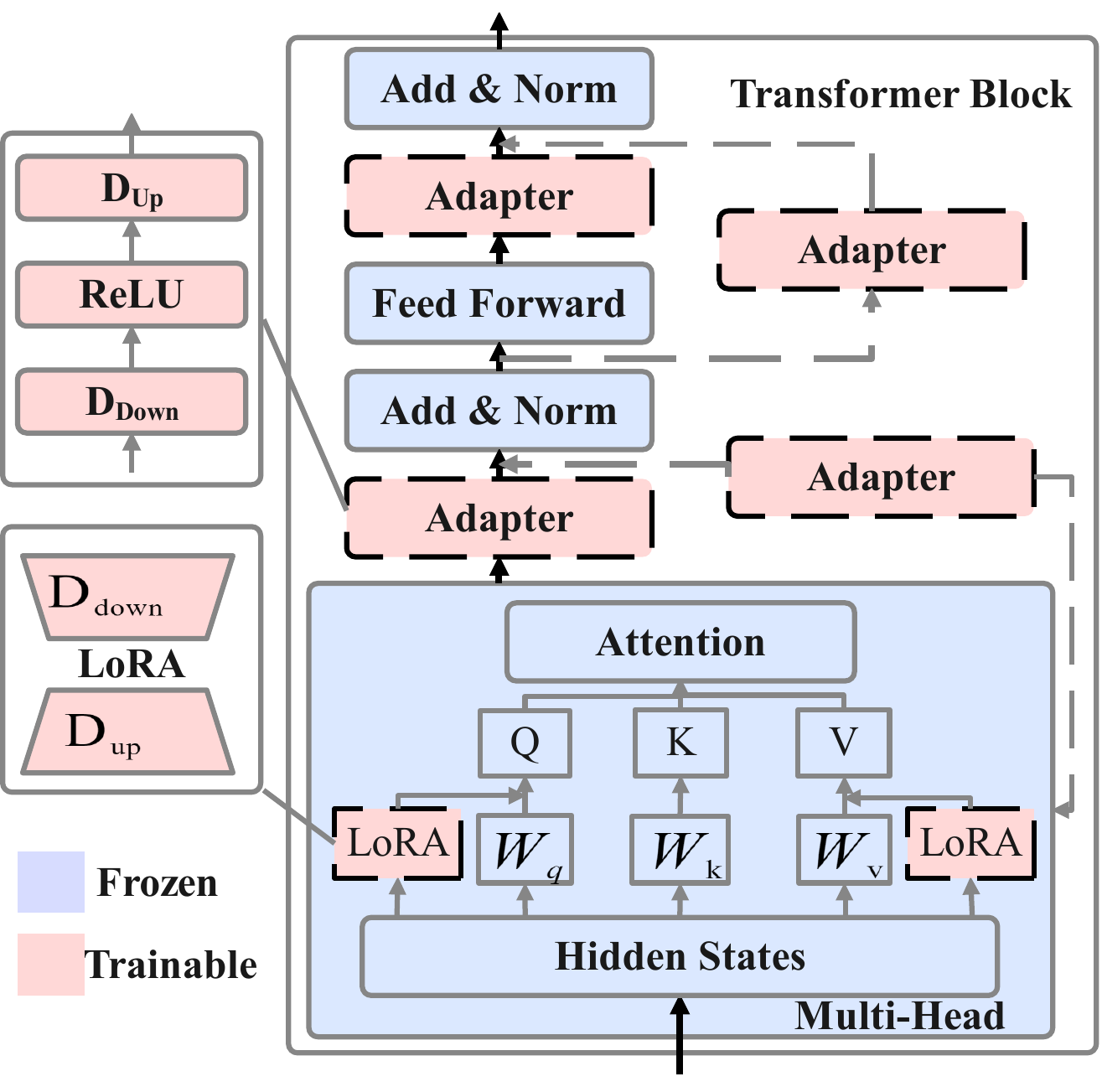}
\caption{Illustration of our presented SU-SAM framework, a simple and unified framework for adapting SAM in underpeformed scenes. SU-SAM brings slight trainable parameters, \emph{e.g.,} adapters and LoRA, to different positions of the Transformer backbone in various manners.} 
\label{Transformer}
\vspace{-1mm}
\end{figure}

To address this issue, a natural question is can we design a simple and unified method that can easily and efficiently fine-tune the SAM model in underperformed scenes?
In this paper, we use the term ``underperformed scene'' to refer to challenging scenarios and tasks where SAM’s performance is relatively unsatisfactory.
We aim to enhance SAM's performance using parameter-efficient techniques while maintaining excellent generalization capabilities on various downstream tasks. 
To this end, we present a simple and unified framework, namely SU-SAM, for improving the Segment Anything Model's adaptability in a parameter-efficient manner to underperformed scenes. In our SU-SAM framework, common parameter-efficient modules (Adapters and LoRAs) are fundamental design components of the overall architecture, unified under a coherent theoretical explanation.
They are conceived as adjustments to specific hidden states within the SAM, facilitating a more coherent alignment of the model's representation space with the feature distribution of individual datasets.
As illustrated in Figure~\ref{Transformer}, the red portion inside the dashed box represents the freely selectable modules and the blue portion represents the original SAM structure with frozen parameters. 
\textit{Note that our SU-SAM removes all adapter-unrelated operations}, such as the carefully designed data augmentations, additional auxiliary input information, and post-processing of outputs. 
This makes our method simple, universal, and data-agnostic that can be applied to any segmentation task dataset. 
Besides, four variants of SU-SAM are introduced to explore the optimal Parameter-Efficient design. Each variant encompasses sequential, parallel, and mixed structures. 

We conduct extensive experiments and analyses on multiple tasks and datasets, including COD10K \cite{COD}, CHAMELEON \cite{skurowski2018animal} and CAMO \cite{le2019anabranch} for camouflage object detection, ISIC \cite{codella2018skin} and ColonDBC \cite{tajbakhsh2015automated} for medical image segmentation, DUTS \cite{DUTS} for salient object segmentation, SBU \cite{SBU} for shadow masking, CDS2K \cite{CSU23} for defect segmentation, and DIS-TE4 \cite{DIS} for complex object shapes. 
These experimental results demonstrate that our SU-SAM achieves competitive or even superior accuracy to state-of-the-art methods. Then, we delved into a detailed discussion and in-depth analyses regarding the parameter settings of each module within SU-SAM. Finally, benefiting from the data-agnostic design, we propose a generalized version of SU-SAM. Unlike previous efforts, this generalized model is trained on multiple datasets simultaneously and consistently delivers competitive performance across all datasets. 

In summary, our main contributions are three-fold: 
\begin{itemize}
\item Without bells and whistles,
we propose a simple and unified framework, SU-SAM, that significantly improves SAM's adaptability toward underperformed scenes without any task-specific design and meanwhile, maintains excellent zero-shot generalizability. Besides, we propose four variants of SU-SAM, \emph{i.e.,} series, parallel, mixed, and LoRA structures.  

\item We provide in-depth analyses and thoughtful insights
that the manner of incorporating parameter-efficient design significantly affects the final performance of SAM-like models. The mixed structure has demonstrated impressive performance on almost all datasets. In addition, the parallel adapter achieves comparable accuracy on multiple datasets. 

\item Extensive experiments with analyses on nine datasets and six segmentation tasks demonstrate the efficacy and efficiency of SU-SAM. It is remarkably simple and flexible to 
deploy our presented SU-SAM in various scenarios and tasks of SAM-like models.
\end{itemize}

\begin{table*}[ht]
\caption{Illustration of the design elements and variants of our SU-SAM.  
}
\label{Methods}
\centering
\begin{center}
\resizebox{1.0\textwidth}{!}{%
\vspace{-1mm}
\begin{tabular}{l l l l l}
\toprule[0.2em]
Method  & $\Delta \boldsymbol{h}$ & Insertion Form & Modified Location & Final Function  \\
\midrule[0.2em]
\multicolumn{5}{c}{\textbf{Design Elements}} \\ 
LoRA &$\boldsymbol{x} \boldsymbol{W}_{\text {down }} \boldsymbol{W}_{\text {up }}$ & parallel & attn key/val & $\boldsymbol{h} \leftarrow \boldsymbol{h}+s \cdot \Delta\boldsymbol{h}$  \\
\multirow{2}{*}{Adapter} &$f\left(\boldsymbol{h} \boldsymbol{W}_{\text {down }}\right) \boldsymbol{W}_{\text {up }}$  & sequential & ffn/attn & $\boldsymbol{h} \leftarrow \boldsymbol{h}+ \Delta\boldsymbol{h}$ \\
 &$f\left(\boldsymbol{h} \boldsymbol{W}_{\text {down }}\right) \boldsymbol{W}_{\text {up }}+\boldsymbol{x}$  & parallel & ffn/attn & $\boldsymbol{h} \leftarrow \boldsymbol{h}+\Delta\boldsymbol{h}$  \\
\hline
\multicolumn{5}{c}{\textbf{Proposed Variants}} \\ 
Series SU-SAM &$f\left(\boldsymbol{h} \boldsymbol{W}_{\text {down }}\right) \boldsymbol{W}_{\text {up }}$ & sequential& ffn/attn & $\boldsymbol{h} \leftarrow \boldsymbol{h}+ \Delta\boldsymbol{h}$ \\
Parallel SU-SAM &$f\left(\boldsymbol{h} \boldsymbol{W}_{\text {down }}\right) \boldsymbol{W}_{\text {up }}+\boldsymbol{x}$& parallel & ffn/attn & $\boldsymbol{h} \leftarrow \boldsymbol{h}+ \Delta\boldsymbol{h}$ \\
Mix SU-SAM &$f\left(\boldsymbol{h} \boldsymbol{W}_{\text {down }}\right) \boldsymbol{W}_{\text {up }}+\boldsymbol{x}$  & both & ffn/attn & $\boldsymbol{h} \leftarrow \boldsymbol{h}+ \Delta\boldsymbol{h}$  \\
LoRA SU-SAM &$\boldsymbol{x} \boldsymbol{W}_{\text {down }} \boldsymbol{W}_{\text {up }}$  & parallel & attn query/val & $\boldsymbol{h} \leftarrow \boldsymbol{h}+s \cdot \Delta\boldsymbol{h}$  \\
\bottomrule[0.2em]
\end{tabular}}
\end{center}
\end{table*}

%% file: latex/2_related_work.tex
\section{Related Work}
\label{sec:related_work}

\subsection{Vision Segmentation Task}
Image segmentation is a fundamental task in computer vision \cite{zhao2018icnet,xiao2018unified,kirillov2019panoptic,wu2019fastfcn,cheng2021per,zhao2018psanet}. It involves dividing an image into different regions or segments. Over the years, researchers have proposed various approaches. FCN \cite{he2019fully} by Long et al. revolutionized vision segmentation with end-to-end pixel-wise predictions. U-Net \cite{ronneberger2015unet} by Ronneberger et al. is a popular architecture for biomedical image segmentation. It uses a U-shaped network with skip connections to capture local and global information effectively. 
Mask R-CNN \cite{he2017mask} by He et al. extends the Faster R-CNN framework for object detection by adding a branch for pixel-level mask predictions. DeepLab models \cite{segmenter2021,deeplabv3+,DeeplabV22017} proposed atrous convolutions and spatial pyramid pooling to improve segment accuracy. In this study, we evaluate SU-SAM in diverse real-world segmentation applications, including natural images, manufacturing, remote sensing, and healthcare.

\subsection{Foundation Models}
Foundation models are a new paradigm in artificial intelligence, based on training large neural networks on massive amounts of data using self-supervised learning techniques \cite{bommasani2021opportunities,wang2023large,liang2022foundations}. They learn general representations that can be transferred to different domains and applications. In the field of NLP, models like BERT \cite{devlin2018bert}, T5 \cite{raffel2020exploring}, and GPT \cite{brown2020language} have demonstrated impressive performance on a wide range of tasks. Similarly, in CV, foundation models aim to learn universal visual representations from large-scale image-text data. These models, Vision-Language Models (VLM) (CLIP~\cite{CLIP} and DALL-E~\cite{ramesh2022hierarchical}) combine computer vision and natural language processing to understand and generate descriptions or analyze visual content using textual and visual information. Masked Image Modeling~\cite{xie2022simmim,liu2022swin} (MIM) refers to a technique where parts of an image are masked during training to encourage a model to learn contextual information and complete missing regions. SAM~\cite{SAM} is a model designed for segmenting objects or areas in images, offering precise segmentation capabilities. Previous works mainly focused on improving the accuracy of vision foundation models to specific downstream tasks. In contrast, our approach aims to propose a general solution that enhances the fine-tuning accuracy of vision foundation models across almost all downstream tasks.

\subsection{Parameter-Efficient Fine-tuning}

Parameter-efficient fine-tuning (PEFT) is a technique widely used for both NLP and CV domains. It allows for efficient model adaptation without extensively modifying the entire network. Two popular methods for implementing this technique are SU-SAM and LoRA.

In NLP, Houlsby et al. introduced Adapter-BERT \cite{houlsby2019parameter}, which utilizes lightweight adapter modules to learn task-specific parameters while preserving the pre-trained weights. Pfeiffer et al. proposed Sentence-BERT with adapters \cite{Geigle2021sentence}, which applies adapters to encode sentence embeddings efficiently.
In CV, Rebuffi et al. proposed RotNetAdapters \cite{Sylvestre2017Learning}, which leverages adapters to predict image rotations and enable self-supervised learning. Li et al. introduced Vision-AdapterNet \cite{Alon2018AdapterNet}, which integrates adapter modules to adapt pre-trained models for image classification.

LoRA \cite{Edward2022LoRA} (Learn Once, Reuse Anywhere) is another approach for parameter-efficient fine-tuning. In NLP, Wang et al. proposed LoRA for BERT \cite{Edward2022LoRA}, which introduces a learnable sparse routing mechanism to share task-specific information across multiple tasks. In CV, \cite{Edward2022LoRA} introduced LoRA for Vision Transformers, enabling efficient adaptation of pre-trained models for various downstream tasks. Previous works \cite{wu2023medical,zhang2023customized,chen2023sam} typically combine PEFT with other task-specific optimizations for specific domain datasets, making it challenging to truly understand the impact of PEFT design itself on the model. We aim to conduct an in-depth analysis of various PEFT methods and identify straightforward approaches that can better leverage their strengths

%% file: latex/3_method.tex
\section{Method}
\label{sec:method}
\subsection{Preliminary}

\textbf{Transformer and SAM:}
Transformer\cite{Vaswani2017Attention}: is composed of several stacked blocks, where each block contains two types of sub-module: multi-head self-attention and a fully connected feed-forward network (FFN). 
For queries $Q\in \mathbb{R}^{n \times d_k}$ and key-value pairs $K \in \mathbb{R}^{m \times d_k}, V \in \mathbb{R}^{m \times d_v}$, the attention maps them as follows:
\begin{equation}
\operatorname{Attn}(\boldsymbol{Q}, \boldsymbol{K}, \boldsymbol{V})=\operatorname{softmax}\left(\frac{\boldsymbol{Q} \boldsymbol{K}^T}{\sqrt{d_k}}\right) \boldsymbol{V}
\end{equation}
where $n$ and $m$ are the numbers of queries and key-value pairs, respectively, and $d_k$ is the dimension of the tokens. FFN, which consists of two linear transformations with ReLU activation function.

SAM\cite{SAM} consists of an image encoder, a prompt encoder, and a lightweight mask decoder. In this work, we mainly focus on the image encoder module, which is a typical transformer architecture. 

\noindent\textbf{Adapter and LoRA:}
\label{Unified}
Inspired by the work\cite{yu2022towards} in the NLP domain, we introduce the basic designs of Adapter and LoRA in CV domain and attempt to derive a unified formulation to establish their connections. 
Our technical discussion is based on the abstraction of methods \cite{chen2023sam}, \cite{wu2023medical}, and \cite{zhang2023customized}, disregarding their designs other than Adapter and LoRA. 
In our baseline configuration, apart from the additional trainable parameters, the parameters of SAM itself are frozen.
The usage of symbols follows \cite{Edward2022LoRA} and \cite{yu2022towards}.

The adapter-related method refers to the insertion of small modules between the Transformer blocks. As shown in Figure \ref{Transformer}, a general adapter module consists of a down-projection layer $W_{down}\in\mathbb{R}^{d \times r}$, a non-linear activation layer $f(\cdot)$ (typically using ReLU function), and an up-projection layer $W_{up}\in\mathbb{R}^{r \times d}$, Here, ``$r$" represents the bottleneck dimension that is significantly smaller than the input dimension. Conventionally (as in \cite{chen2023sam}), adapter operations are applied sequentially, with one adapter module inserted after the attention and another after the FFN. These adapters are surrounded by a residual connection. For modules with weight matrix $W_0 \in \mathbb{R}^{d \times k}$ and $h=W_0 x$, the final form is:
\begin{equation}
\boldsymbol{h} \leftarrow \boldsymbol{h}+f\left(\boldsymbol{h} \boldsymbol{W}_{\text {down }}\right) \boldsymbol{W}_{\text {up }}
\end{equation}
In our work, as shown in Figure \ref{Transformer}, we also explore the parallel implementation of adapters. The adapter operates in parallel with the existing attention layer and FFN, as follows:
\begin{equation}
\boldsymbol{h} \leftarrow \boldsymbol{h}+ f\left(\boldsymbol{x} \boldsymbol{W}_{\text {down }}\right) \boldsymbol{W}_{\text {up }}+x
\end{equation}

LoRA \cite{Edward2022LoRA} hypothesizes the updates to the weights also have a low ``intrinsic rank'' during adaptation. For a module with weight matrix $W_0 \in \mathbb{R}^{d \times k}$, LoRA constrains its update by representing the latter with a low-rank decomposition $ W_0+\Delta W=W_0+B A, \text { where } B \in \mathbb{R}^{d \times r}, A \in \mathbb{R}^{r \times k} $, and the rank $r\ll \min (d, k)$. We consider $A$ as the down-projection layer $W_{down}$ and  $B$ as the up-projection layer $W_{up}$. With $h=W_0 x$, it leads to a final form: 
\begin{equation}
\boldsymbol{h} \leftarrow \boldsymbol{h}+s \cdot \boldsymbol{x} \boldsymbol{W}_{\text {down }} \boldsymbol{W}_{\text {up }}
\end{equation}
and $s$ is the hyperparameter to control the effect of LoRA.

\subsection{SU-SAM with Four Variants}
For the similarity between Adapter and LoRA, we propose a unified framework to consolidate existing related works, called SU-SAM. We assume an initial parameter matrix, denoted as $W_0$, and original output $h=W_0 x$. The parameter matrix that best matches a specific dataset is denoted as $W_f$, with the final output $h_f=W_1 x$. Specifically, we cast Adapter and LoRA as learning a residual vector $\Delta \boldsymbol{h}$ between $h$ and $h_f$. 
Throughout the fine-tuning process, $\boldsymbol{h}$ and $\boldsymbol{x}$ remain unchanged, and only the residual vector $\Delta \boldsymbol{h}$ is modified. As shown in Table \ref{Methods}, to better describe this modification process, we define a set of dimensions that define different methods as a combination of instances along these dimensions. 

\noindent \textbf{Functional Form} is the specific function that computes $\Delta \boldsymbol{h}$. 
We conducted a detailed discussion above and abstracted the Adapter and LoRA into a unified structure of \textit{Down-projection-Nonlinear-Up-projection}. Usually, we use ReLU as the Nonlinear function.

\noindent \textbf{Insertion Form} is how the added module is inserted into the network. We divide it into two types: sequential and parallel insertion.

\noindent \textbf{Modified Location} represents the specific module location within the Transformer where the insertion occurs.

\noindent \textbf{Final Function} is the final form after the insertion.

\begin{figure*}[t]
\centering
\includegraphics[width=1.0\textwidth]{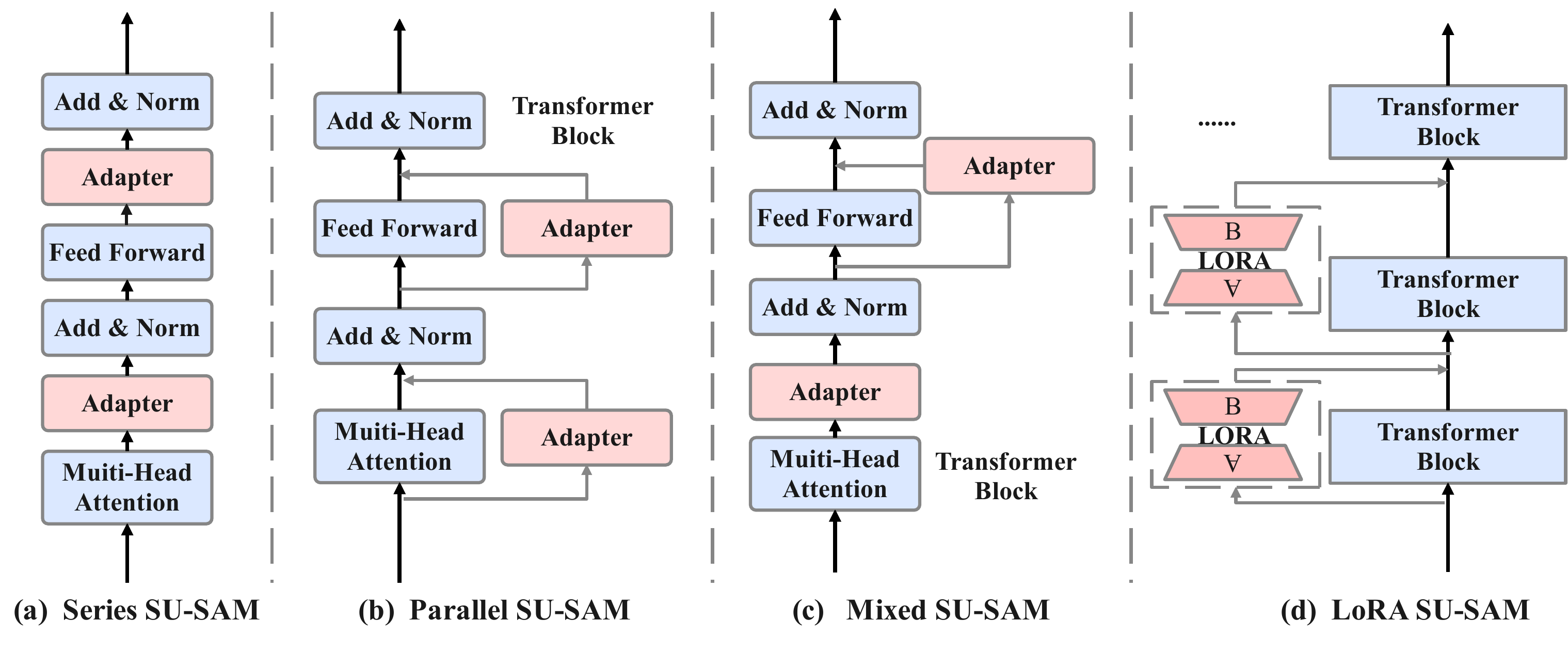} 
\vspace{-2mm}
\caption{Illustration of 
the model architectures of four different variants, \emph{i.e.,} Series SU-SAM, Parallel SU-SAM, Mixed SU-SAM, LoRA SU-SAM. In this figure, some SAM modules are omitted, and we solely focus on modified components. }
\label{Variants}
\end{figure*}

 Based on the proposed SU-SAM and design elements, we derive four new variants, as shown in Figure \ref{Variants}. Detailed comparisons between these variants and the original design dimensions are provided in Table \ref{Methods}.
 (1) \textbf{Series SU-SAM}: Two adapter modules are inserted between the attention and FFN layers sequentially. (2) \textbf{Parallel SU-SAM}: Similar to (1), we insert two adapter modules, but they are in parallel with the original structure. (3) \textbf{Mixed SU-SAM}: It is a hybrid structure where a serial Adapter is used in the attention layer and a parallel Adapter is used in the FFN layer. (4) \textbf{LoRA SU-SAM}: We apply a LoRA structure with a scalar hyperparameter $s$ to the query and value projection layers in the attention module. 

By proposing four variants of SU-SAM, we aim to find a way that can easily and efficiently fine-tune the SAM model with parameter-efficient techniques while maintaining excellent generalizability toward various downstream tasks. All of these SU-SAM variants do not require any task-specific designs and aim to significantly improve the adaptability of SAM-like models toward underperformed scenes. 
In the following sections, we will conduct in-depth discussions and research.

\subsection{Discussion with Other Arts}

As mentioned in the Introduction section, many optimization methods based on SAM  combine parameter-efficient techniques with task-specific modules, which restricts their ability to generalize to datasets from other domains.
In this section, we discuss the difference between our SU-SAM framework with prior arts that belong to other adapter-based SAM methods. 

As shown in Figure \ref{teaser}, the blue color blocks indicate the frozen modules that cannot be trained, and the pink color blocks represent the structures that are trainable during fine-tuning. Besides, we use the yellow color to represent task-specific designs in other methods. It is clear that, compared to previous works (e.g, MSA \cite{wu2023medical}, SAMed \cite{zhang2023customized}, SAM-Adapter \cite{chen2023sam}), we simplify the whole pipeline. The  detailed differences lie in the following aspects:
\begin{itemize}
\item \textbf{We do not require additional dataset-specific pretraining.} Instead, we keep the original parameters of SAM unchanged and freeze them during the fine-tuning.
\item \textbf{We do not need additional data preprocessing.} There is no need for designs similar to SAM-Adapter, which incorporates high-frequency information of the original images as additional input patch embeddings.
\item \textbf{We maintain the original structure of SAM without any modifications}. We do not introduce additional multi-scale or multi-branch designs.
\item \textbf{We do not require additional data post-processing.} Our input and output remain consistent with the original SAM definition, without employing extra post-processing techniques to enhance model accuracy.
\end{itemize}

%% file: latex/4_experiment.tex
\section{Experiments}
\label{sec:exp}

\subsection{Benchmark Datasets of Different Tasks}
Following results in \cite{ji2023segment}, we verify SU-SAM on the datasets where SAM underperformed. We prove that our SU-SAM could significantly improve the performance of all of the datasets. The datasets encompass various downstream tasks, including salient object segmentation \cite{borji2019salient1}, complex object shapes segmentation \cite{DIS}, camouflaged object segmentation \cite{COD}, shadow detection \cite{shadow}, surface defect segmentation \cite{he2019fully}, and medical field segmentation \cite{tajbakhsh2015automated}.

\noindent \textbf{Salient object segmentation } \cite{borji2019salient1,zhou2021rgb,ge2021wgi} aims to identify and extract the most visually prominent objects or regions in an image. Accurate salient object segmentation enables the automatic highlighting of important objects, leading to improved image understanding and various downstream applications in CV. \textbf{DUTS} \cite{DUTS} is currently the largest salient object segmentation dataset, consisting of 10,553 training images and 5,019 test images.

\noindent \textbf{Complex object shapes} \cite{DIS} refer to objects that exhibit intricate or intricate geometries, often characterized by irregular contours, multiple components, or intricate structures. Complex object shapes pose challenges for vision tasks, including object recognition, segmentation, and understanding, as they require advanced algorithms to accurately capture their intricate characteristics and accurately delineate their boundaries. \textbf{DIS-TE4} \cite{DIS} provides 500 highly accurate pixel-level masks for images with complex object shapes. This can evaluate SAM’s ability to perceive boundary details.

\begin{figure*}[h!t]
\centering
\includegraphics[width=1.0\textwidth]{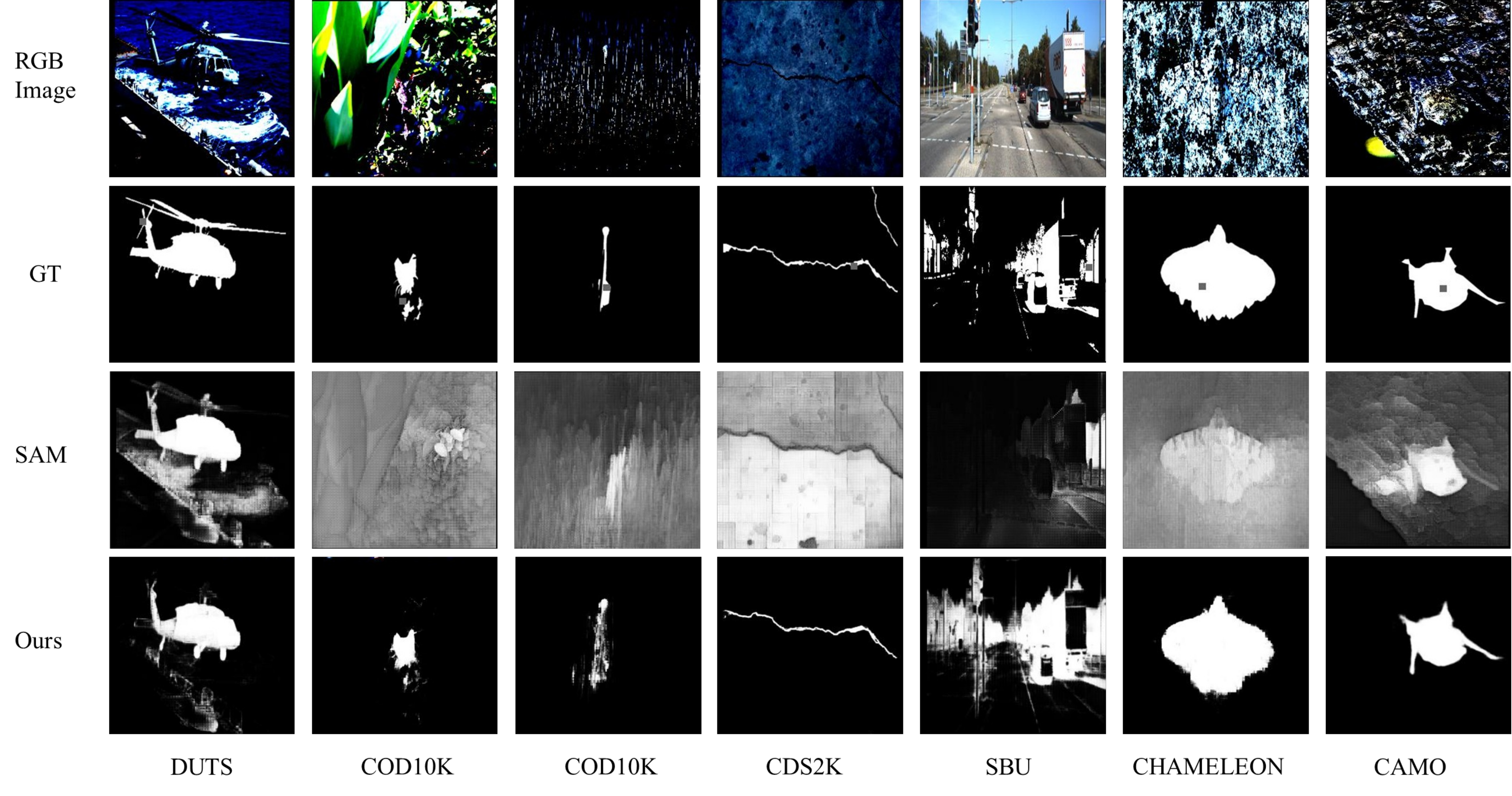} 
\vspace{-2mm}
\caption{
\textbf{Visualization results.} We randomly selected images from the test sets of several datasets (COD10K \cite{COD}, SBU \cite{SBU}, DUTS \cite{DUTS}, CHAMELEON~\cite{skurowski2018animal}, CAMO~\cite{le2019anabranch}).SU-SAM significantly improves the performance of the original SAM.}
\label{visualize}
\vspace{2mm}
\end{figure*}

\noindent \textbf{Camouflaged object detection}  \cite{COD,zhou2023specificity,fan2021re} is a task that aims to identify and localize objects that are visually disguised or blended into their surrounding environments. This task poses significant challenges due to camouflage patterns' complex and diverse nature, which often involve color, texture, and shape adaptations. \textbf{COD10K} \cite{COD} is a substantial dataset specifically designed for camouflaged object segmentation, featuring 10,000 images characterized by similar foreground objects and surroundings. \textbf{CAMO} \cite{le2019anabranch} dataset includes 76 images collected from the Internet for testing.
\textbf{CHAMELEON} \cite{skurowski2018animal} consists of 1250 images (1000 images for the training set and 250 images for the testing set).

\noindent \textbf{Shadow detection } \cite{SBU} is a task that involves identifying and distinguishing shadows from the corresponding objects or regions in an image. Shadows can significantly affect image analysis and understanding, as they introduce variations in color, texture, and shape. The goal of shadow detection is to accurately segment and classify shadow regions, enabling applications such as object recognition, scene understanding, and image-based lighting estimation. \textbf{SBU} \cite{SBU} provides carefully annotated shadow masks for 700 images, which is widely used in shadow detection tasks.

\noindent \textbf{Surface defect segmentation } \cite{he2019fully} aims at identifying and delineating defective regions in an image. It involves automatically detecting and segmenting areas that deviate from the normal appearance, such as scratches, cracks, stains, or other imperfections. Defect segmentation is essential in quality control and inspection processes across various industries, including manufacturing, automotive, electronics, and product packaging. \textbf{CDS2K} \cite{CSU23} is a comprehensive concealed defect segmentation dataset, compiled from a variety of established industrial defect repositories. It comprises 2,492 samples, consisting of 1,330 positive and 1,162 negative.

\noindent \textbf{Medical Segmentation}: \textbf{ISIC} \cite{codella2018skin} and \textbf{ColonDB} \cite{ColonDB} are used to verify the generalization capability of SAM in the medical field. The ISIC dataset consists of high-resolution images of skin lesions, along with corresponding ground truth annotations for lesion segmentation and classification. ColonDB consists of 380 images.

\subsection{Implementation Details}
To ensure a fair comparison, we utilize the same codebase and maintain consistency in training hyperparameter settings unrelated to the methods. We refer to \cite{wu2023medical}, \cite{chen2023sam}, and \cite{zhang2023customized} for guidance. The modifications are solely made to the design dimensions mentioned in Section \ref{3}. We uniformly resized the images to $1024\times1024$ without additional data preprocessing. The structure of all inserted adapters was fixed as shown in Figure \ref{Transformer}: a downsampling layer, a ReLU activation function, and an upsampling layer. The hidden dimension was set to 1/4 of the input dimension. All inserted LoRAs were uniformly set with a rank of 4. We used the ViT-B version of SAM and trained with the BCE loss (Binary Cross-Entropy loss). AdamW \cite{Loshchilov2019Decoupled} optimizer was used for all experiments with an initial learning rate of 1e-4. The StepLR scheduler was employed with a decay step of 10 and a gamma value of 0.5. Fine-tuning was performed for 100 epochs on all datasets. All experiments were conducted on a single NVIDIA GTX 4090 GPU for fair comparisons.

\subsection{Evaluation metrics} 
In the experiments, we use the widely used Mean Absolute Error (MAE) and Average Precision (AP) for evaluation. A lower MAE score and a higher AP score indicate better model performance. 

\begin{table*}[!ht]
\vspace{.3cm}
 \caption{ Quantitative results on applications of (a) salient object segmentation in \textit{common scenes}, (b) salient object segmentation with \textit{highly-accurate details}, (c) shadow detection, (d) concealed industrial defect detection (e) and (f) medical polyp lesion segmentation, (g) (h) and (i) camouflaged object segmentation. $\Delta$ shows the performance gaps between SAM and our proposed SU-SAM. We highlighted the best performance values on each dataset.}
    \label{Results}
  \centering
    \begin{subtable}{.26\linewidth}
    \centering
    \caption{DUTS~\cite{DUTS}. }
\resizebox{!}{1.65 cm}{    
\begin{tabular}{c|c|c|c}
    \hline
    \multicolumn{2}{c|}{Model} & Backbone  & MAE \\
    \hline
    \multicolumn{2}{c|}{VST$_{21}$~\cite{VST}} & T2T-ViTt& 0.037 \\
    \multicolumn{2}{c|}{ICONet$_{22}$~\cite{zhuge2022salient}} & ResNet50  & 0.037 \\
    \multicolumn{2}{c|}{SAM~\cite{SAM}}  & ViT-B  & 0.298 \\
    \hline
    \multirow{8}[2]{*}{SU-SAM} & Para & \multirow{2}[2]{*}{ViT-B}  &   0.055 \\
          & \small{$\Delta$\textit{diff}}	& 	&	\small{\textcolor[rgb]{ .459,  .443,  .443}{$\uparrow$24.3\%}} \\ \cline{2-4}
        & Series 		& \multirow{2}[2]{*}{ViT-B}		&	0.148   \\
          & \small{$\Delta$\textit{diff}}	& 	&	  \small{\textcolor[rgb]{ .459,  .443,  .443}{$\uparrow$15.0\%}} \\ \cline{2-4}
      & Mix 		& \multirow{2}[2]{*}{ViT-B}		&	\textbf{0.029} \\
          & \small{$\Delta$\textit{diff}}	&  	&	   \small{\textcolor[rgb]{ .459,  .443,  .443}{$\uparrow$26.9\%}}   \\ \cline{2-4}
         & LoRA		& \multirow{2}[2]{*}{ViT-B}		&	 0.079 \\
          & \small{$\Delta$\textit{diff}}		&        & \small{\textcolor[rgb]{ .459,  .443,  .443}{$\uparrow$21.9\%}}   \\ \cline{2-4}
    \hline
\end{tabular}}
  \end{subtable}
  \hspace{\fill}
    \begin{subtable}{.26\linewidth}
    \centering
    \caption{DIS-TE4~\cite{DIS}. }
\resizebox{!}{1.65 cm}{    
\begin{tabular}{c|c|c|c}
    \hline
    \multicolumn{2}{c|}{Model} & Backbone  & MAE \\
    \hline
    \multicolumn{2}{c|}{Gate$_{20}$~\cite{zhao2020suppress}} & ResNet50	 	& 0.109 \\
    \multicolumn{2}{c|}{IS-Net~\cite{DIS}} & SwinB		& 0.072 \\
    \multicolumn{2}{c|}{SAM~\cite{SAM}} 	& ViT-B & 0.373 \\
    \hline
    \multirow{8}[2]{*}{SU-SAM} & Para  & \multirow{2}[2]{*}{ViT-B}   & 0.080 \\
          & \small{$\Delta$\textit{diff}}		&  & \small{\textcolor[rgb]{ .459,  .443,  .443}{$\uparrow$29.3\%}} \\ \cline{2-4}
        & Series 		& \multirow{2}[2]{*}{ViT-B}		&	  0.191 \\
          & \small{$\Delta$\textit{diff}}		&	  &  \small{\textcolor[rgb]{ .459,  .443,  .443}{$\uparrow$18.2\%}} \\ \cline{2-4}
      & Mix 			& \multirow{2}[2]{*}{ViT-B}	&	 \textbf{0.049} \\
          & \small{$\Delta$\textit{diff}}		  &    & \small{\textcolor[rgb]{ .459,  .443,  .443}{$\uparrow$32.4\%}}   \\ \cline{2-4}
         & LoRA	& \multirow{2}[2]{*}{ViT-B}			  & 0.117 \\
          & \small{$\Delta$\textit{diff}}		&      & \small{\textcolor[rgb]{ .459,  .443,  .443}{$\uparrow$25.6\%}}   \\ \cline{2-4}
    \hline
    \end{tabular}}
  \end{subtable}
  \hspace{\fill}
    \begin{subtable}{.26\linewidth}
    \centering
   \caption{SBU~\cite{SBU}.}
\resizebox{!}{1.65 cm}{    
\begin{tabular}{c|c |c|c}
    \hline
    \multicolumn{2}{c|}{Model} & Backbone  & MAE  \\
    \hline
    \multicolumn{2}{c|}{DSC$_{18}$~\cite{hu2018direction}} &VGG-16  & \textbf{0.032} \\
    \multicolumn{2}{c|}{DSDNet$_{19}$~\cite{zheng2019distraction}} &ResNext  & 0.036 \\
    \multicolumn{2}{c|}{SAM~\cite{SAM}} & ViT-B &   0.496  \\
    \hline
    \multirow{8}[2]{*}{SU-SAM} & Para & \multirow{2}[2]{*}{ViT-B}  & 0.234 \\
          & \small{$\Delta$\textit{diff}}		&   & \small{\textcolor[rgb]{ .459,  .443,  .443}{$\uparrow$26.2\%}} \\ \cline{2-4}
        & Series 		& \multirow{2}[2]{*}{ViT-B}		 & 0.079 \\
          & \small{$\Delta$\textit{diff}}			&  &  \small{\textcolor[rgb]{ .459,  .443,  .443}{$\uparrow$41.7\%}} \\ \cline{2-4}
      & Mix 			& \multirow{2}[2]{*}{ViT-B}	   & 0.070 \\
          & \small{$\Delta$\textit{diff}}		 &    & \small{\textcolor[rgb]{ .459,  .443,  .443}{$\uparrow$42.6\%}}   \\ \cline{2-4}
         & LoRA			& \multirow{2}[2]{*}{ViT-B}	&	0.180  \\
          & \small{$\Delta$\textit{diff}}		&      & \small{\textcolor[rgb]{ .459,  .443,  .443}{$\uparrow$31.6\%}}   \\ \cline{2-4}
    \hline
    \end{tabular}}
  \end{subtable}
    \begin{subtable}{.26\linewidth}
    \centering
    \caption{CDS2K~\cite{CSU23}. }
\resizebox{!}{1.65 cm}{    
\begin{tabular}{c|c|c|c}
    \hline
    \multicolumn{2}{c|}{Model} & Backbone  & MAE \\
    \hline
    \multicolumn{2}{c|}{SINetV2$_{22}$~\cite{fan2021concealed}} & Res2Net50   	& \textbf{0.102} \\
    \multicolumn{2}{c|}{HitNet$_{23}$~\cite{HiNet}} 	&	PVTv2-B2  	& 0.118  \\
    \multicolumn{2}{c|}{SAM~\cite{SAM}} & ViT-B &	 0.611 \\
    \hline
    \multirow{8}[2]{*}{SU-SAM} & Para & \multirow{2}[2]{*}{ViT-B} & 0.137 \\
          & \small{$\Delta$\textit{diff}}		&  & \small{\textcolor[rgb]{ .459,  .443,  .443}{$\uparrow$47.4\%}} \\ \cline{2-4}
        & Series 		& \multirow{2}[2]{*}{ViT-B}		&	0.123   \\
          & \small{$\Delta$\textit{diff}}		&	  &  \small{\textcolor[rgb]{ .459,  .443,  .443}{$\uparrow$48.8\%}} \\ \cline{2-4}
      & Mix 			& \multirow{2}[2]{*}{ViT-B}	&	\textbf{0.102} \\
          & \small{$\Delta$\textit{diff}}		&      & \small{\textcolor[rgb]{ .459,  .443,  .443}{$\uparrow$50.9\%}}   \\ \cline{2-4}
         & LoRA			& \multirow{2}[2]{*}{ViT-B}	&	0.165  \\
          & \small{$\Delta$\textit{diff}}		  &     & \small{\textcolor[rgb]{ .459,  .443,  .443}{$\uparrow$44.6\%}}   \\ \cline{2-4}
    \hline
    \end{tabular}}
  \end{subtable}
  \hspace{\fill}
\begin{subtable}{.26\linewidth}
    \centering
   \caption{ISIC~\cite{codella2018skin}.}
\resizebox{!}{1.65 cm}{    
\begin{tabular}{c|c| c|c}
    \hline
    \multicolumn{2}{c|}{Model} & Backbone  & MAE \\
    \hline
    \multicolumn{2}{c|}{CASS~\cite{singh2022cass}} & CNN+TF  & 0.086 \\
    \multicolumn{2}{c|}{DINO~\cite{zhang2022dino}} & ViT-B & 0.081 \\
    \multicolumn{2}{c|}{SAM~\cite{SAM} } &  ViT-B & 0.419 \\
    \hline
    \multirow{8}[2]{*}{SU-SAM} & Para & \multirow{2}[2]{*}{ViT-B} & 0.062 \\
          & \small{$\Delta$\textit{diff}}		&  & \small{\textcolor[rgb]{ .459,  .443,  .443}{$\uparrow$35.7\%}} \\ \cline{2-4}
        & Series 			& \multirow{2}[2]{*}{ViT-B}	&	 \textbf{0.044} \\
          & \small{$\Delta$\textit{diff}}		  & & \small{\textcolor[rgb]{ .459,  .443,  .443}{$\uparrow$37.5\%}} \\ \cline{2-4}
      & Mix 			& \multirow{2}[2]{*}{ViT-B}	&	 0.054 \\
          & \small{$\Delta$\textit{diff}}		   &    & \small{\textcolor[rgb]{ .459,  .443,  .443}{$\uparrow$36.5\%}}   \\ \cline{2-4}
         & LoRA			& \multirow{2}[2]{*}{ViT-B}	&	0.087  \\
          & \small{$\Delta$\textit{diff}}		&       & \small{\textcolor[rgb]{ .459,  .443,  .443}{$\uparrow$33.2\%}}   \\ \cline{2-4}
    \hline
    \end{tabular}}
  \end{subtable}
  \hspace{\fill}
    \begin{subtable}{.26\linewidth}
    \centering
   \caption{ColonDB~\cite{tajbakhsh2015automated}.}
\resizebox{!}{1.65 cm}{    
\begin{tabular}{c|c|c|c}
    \hline
    \multicolumn{2}{c|}{Model} & Backbone  & MAE \\
    \hline
    \multicolumn{2}{c|}{FAPNet$_{22}$~\cite{zhou2022feature}} & Res2Net50  & 0.038 \\
    \multicolumn{2}{c|}{CFA-Net$_{23}$~\cite{zhou2023cross}} & Res2Net50 & 0.039 \\
    \multicolumn{2}{c|}{SAM~\cite{SAM}} &  ViT-B & 0.111 \\
    \hline
    \multirow{8}[2]{*}{SU-SAM} & Para & \multirow{2}[2]{*}{ViT-B} & 0.045 \\
          & \small{$\Delta$\textit{diff}}		&  & \small{\textcolor[rgb]{ .459,  .443,  .443}{$\uparrow$6.6\%}} \\ \cline{2-4}
        & Series 			& \multirow{2}[2]{*}{ViT-B}	&	  \textbf{0.031} \\
          & \small{$\Delta$\textit{diff}}		&  &  \small{\textcolor[rgb]{ .459,  .443,  .443}{$\uparrow$8.0\%}} \\ \cline{2-4}
      & Mix 		& \multirow{2}[2]{*}{ViT-B}		&	 0.037 \\
          & \small{$\Delta$\textit{diff}}		&       & \small{\textcolor[rgb]{ .459,  .443,  .443}{$\uparrow$7.4\%}}   \\ \cline{2-4}
         & LoRA		& \multirow{2}[2]{*}{ViT-B}		&	 0.055 \\
          & \small{$\Delta$\textit{diff}}		 &      & \small{\textcolor[rgb]{ .459,  .443,  .443}{$\uparrow$5.5\%}}   \\ \cline{2-4}
    \hline
    \end{tabular}}
  \end{subtable}
  \begin{subtable}{.26\linewidth}
    \centering
    \caption{COD10K~\cite{COD}. }
\resizebox{!}{1.65 cm}{    
\begin{tabular}{c|c|c|c}
    \hline
    \multicolumn{2}{c|}{Model} & Backbone  & MAE  \\
    \hline
    \multicolumn{2}{c|}{SegMaR$_{22}$~\cite{jia2022segment}} 	&ResNet50 & 0.034  \\
    \multicolumn{2}{c|}{SINet~\cite{COD}} 	&ResNet50   & 0.092 \\
    \multicolumn{2}{c|}{SAM~\cite{SAM}} &	ViT-B & 0.217 \\
    \hline
    \multirow{8}[2]{*}{SU-SAM} & Para & \multirow{2}[2]{*}{ViT-B}  & 0.054 \\
          & \small{$\Delta$\textit{diff}}		&  & \small{\textcolor[rgb]{ .459,  .443,  .443}{$\uparrow$16.3\%}} \\ \cline{2-4}
        & Series 		& \multirow{2}[2]{*}{ViT-B}		&	 0.051 \\
          & \small{$\Delta$\textit{diff}}		&	  &  \small{\textcolor[rgb]{ .459,  .443,  .443}{$\uparrow$16.6\%}} \\ \cline{2-4}
      & Mix 		& \multirow{2}[2]{*}{ViT-B}		&	\textbf{0.025}  \\
          & \small{$\Delta$\textit{diff}}		&      & \small{\textcolor[rgb]{ .459,  .443,  .443}{$\uparrow$19.2\%}}   \\ \cline{2-4}
         & LoRA		& \multirow{2}[2]{*}{ViT-B}		&	0.077  \\
          & \small{$\Delta$\textit{diff}}	   &   & \small{\textcolor[rgb]{ .459,  .443,  .443}{$\uparrow$14\%}}   \\ \cline{2-4}
    \hline
    \end{tabular}}
\end{subtable}
\hspace{\fill}
    \begin{subtable}{.26\linewidth}
    \centering
   \caption{CAMO~\cite{le2019anabranch}.}
\resizebox{!}{1.65 cm}{    
\begin{tabular}{c|c|c|c}
    \hline
    \multicolumn{2}{c|}{Model} & Backbone  & MAE \\
    \hline
    \multicolumn{2}{c|}{SINet~\cite{COD}}  & ResNet50 & 0.100 \\
    \multicolumn{2}{c|}{FBNet~\cite{FBNet}} & ResNet50 & 0.081 \\
    \multicolumn{2}{c|}{SAM~\cite{SAM}} & ViT-B & 0.286 \\
    \hline
    \multirow{8}[2]{*}{SU-SAM} & Para & \multirow{2}[2]{*}{ViT-B} & 0.117 \\
          & \small{$\Delta$\textit{diff}}	&	  & \small{\textcolor[rgb]{ .459,  .443,  .443}{$\uparrow$16.9\%}} \\ \cline{2-4}
        & Series 		& \multirow{2}[2]{*}{ViT-B}		& 0.108	 \\
          & \small{$\Delta$\textit{diff}}		&	  &  \small{\textcolor[rgb]{ .459,  .443,  .443}{$\uparrow$17.8\%}} \\ \cline{2-4}
      & Mix 		& \multirow{2}[2]{*}{ViT-B}		&	 \textbf{0.070} \\
          & \small{$\Delta$\textit{diff}}		&      & \small{\textcolor[rgb]{ .459,  .443,  .443}{$\uparrow$21.6\%}}   \\ \cline{2-4}
         & LoRA		& \multirow{2}[2]{*}{ViT-B}		   & 0.145 \\
          & \small{$\Delta$\textit{diff}}	 &    & \small{\textcolor[rgb]{ .459,  .443,  .443}{$\uparrow$14.1\%}}   \\ \cline{2-4}
    \hline
    \end{tabular}}
  \end{subtable}
  \hspace{\fill}
  \begin{subtable}{.26\linewidth}
    \centering
    \caption{CHAMELEON~\cite{skurowski2018animal}. }
\resizebox{!}{1.65 cm}{    
\begin{tabular}{c|c|c|c}
    \hline
    \multicolumn{2}{c|}{Model} & Backbone  & MAE \\
    \hline
    \multicolumn{2}{c|}{SINet~\cite{COD}}  & ResNet50 & 0.440 \\
    \multicolumn{2}{c|}{FBNet~\cite{FBNet}} & ResNet50 & 0.032 \\
    \multicolumn{2}{c|}{SAM~\cite{SAM}}  & ResNet50 & 0.334 \\
    \hline
    \multirow{8}[2]{*}{SU-SAM} & Para & \multirow{2}[2]{*}{ViT-B}& 0.050 \\
          & \small{$\Delta$\textit{diff}}	&	  & \small{\textcolor[rgb]{ .459,  .443,  .443}{$\uparrow$28.4\%}} \\ \cline{2-4}
        & Series 	& \multirow{2}[2]{*}{ViT-B}			&	\textbf{0.033}   \\
          & \small{$\Delta$\textit{diff}}	&	  &  \small{\textcolor[rgb]{ .459,  .443,  .443}{$\uparrow$30.1\%}} \\ \cline{2-4}
      & Mix 	& \multirow{2}[2]{*}{ViT-B}		&	\textbf{0.033} \\
          & \small{$\Delta$\textit{diff}}		&      & \small{\textcolor[rgb]{ .459,  .443,  .443}{$\uparrow$30.1\%}}   \\ \cline{2-4}
         & LoRA		& \multirow{2}[2]{*}{ViT-B}		&  0.055 \\
          & \small{$\Delta$\textit{diff}}		&      & \small{\textcolor[rgb]{ .459,  .443,  .443}{$\uparrow$27.9\%}}   \\ \cline{2-4}
    \hline
    \end{tabular}}
  \end{subtable}
\end{table*}

\subsection{Main Results}
As shown in Table~\ref{Results}, we compare our four variants with the SOTA methods on each dataset, as well as SAM (without finetuning) and adapter methods improved specifically for each dataset, to demonstrate the superiority of our approach. We visualize our segmentation results in Figure~\ref{visualize} that compare SU-SAM with the original SAM. Due to space constraints, we randomly selected a subset of images from nine datasets. It can be observed that SU-SAM significantly improves the accuracy of SAM on underperformed scenes. \textit{For more visualization results, please refer to the supplementary material.}

From Table~\ref{Results}, we observe that our proposed variants significantly improve the accuracy of SAM on all datasets. For some datasets, our methods achieve similar or even surpass SOTA performance (\emph{e.g.,} COD10K, CAMO, CHAMELEON, DIS-TE4). Among the four variants, the Mix SU-SAM generally achieves the best accuracy.  Although adapters are commonly implemented using a sequential structure, our design, and experiments demonstrate that parallel structures (Parallel SU-SAM and LoRA SU-SAM) can achieve superior accuracy compared to sequential structures on several datasets (e.g., DUTS, DIS-TE4). Moreover, the parallel structure allows for a less constrained and more flexible design of the inserted modules, as it is less bounded by the constraints of the original structure. This provides new insights for future researchers in selecting models.

It is obvious that the fine-tuned version SU-SAM outperforms the SAM models as visualized in \ref{visualize}. Next, we compare the performance of our SU-SAM with other SAM-based fine-tuning approaches, as well as strategies involving freezing the encoder and only fine-tuning the decoder, to demonstrate the necessity and effectiveness of our SU-SAM. As shown in Table~\ref{results_supp}, under settings with similar numbers of trainable parameters, our method achieves the best performance.

\begin{table}[h!]
\centering
\caption{Comparisons with other fine-tuning methods.
}
\vspace{1mm}
\resizebox{0.8\linewidth}{!}{
\begin{tabular}{c|c| cccc}
\hline
 Method & BackBone & COD10K & DUTS  & DIS & ISIC \\
\hline
SAM-adapter~\cite{chen2023sam} & SAM & 0.026 & -  & - & -  \\
MSA~\cite{wu2023medical} & SAM &- &  - & -  & 0.049     \\
HQ-SAM~\cite{shen2022high} & SAM & - & -  & 0.053 & -  \\
FT-Decoder& SAM & 0.044 & 0.048 & 0.071 & 0.057 \\
SU-SAM  & SAM & \textbf{0.025} & \textbf{0.029}  & \textbf{0.049} & \textbf{0.044}  \\
\hline
\end{tabular}
}
\label{results_supp}
\end{table}

\subsection{Ablation Study}
In this Section, we delve into the selection of Adapter and LoRA. As shown in Table \ref{tab:ablation}, we conducted ablation experiments on the layer numbers and hidden dimensions of SU-SAM on the ISIC dataset. Our default insertion order is MIX which yields the best accuracy in the main results section. The following conclusions are drawn:

    \textbf{1.} Under the setting of one module, varying the hidden dimension of the Adapter leads to significant parameter changes (From 95.91M to 111.56M) but does not improve accuracy performance (From 0.045 to 0.052). This is due to the limited size of datasets, which hampers effective training (as indicated in (a) in Table~\ref{tab:ablation}). 
    
    \textbf{2.}  With keeping the original hidden dimension equal to 0.25 unchanged, increasing the stack number of Adapters significantly increases the parameter count (From 101.26M to 112.46M). However, similar to the first conclusion, no accuracy improvement is observed on small downstream tasks (as shown in (b) in Table~\ref{tab:ablation}).
    
    \textbf{3.}  Considering the low-rank setting of LoRA, modifications to its parameter settings (such as modifying hidden dimension $r$ or increasing the stack number) result in minimal changes in parameter increase (<1M). However, these changes have a negligible impact on model accuracy (as shown in (c) and (d) in Table~\ref{tab:ablation}).

\begin{table}[ht] 
\centering
\caption{ More discussion about adapter and LoRA. Acc.: Accuracy. The evaluation metric is MAE. }
\label{tab:ablation}
\begin{subtable}{0.5\linewidth}
\centering
\caption{Adapter's hidden dimension}
\begin{tabular}{c|ccc}
\hline
Rate & 1/8 & 1/4 & 1/2 \\
\hline
Acc. & 0.052  & 0.045  & 0.052 \\
Size &  95.91 & 101.26 & 111.86 \\
\hline
\end{tabular}
\label{tab:dimention_adapter}
\end{subtable} 
\begin{subtable}{0.49\linewidth}
\centering
\caption{The number of Adapter}
\begin{tabular}{c|ccc}
\hline
 & 1 & 2 & 3 \\
\hline
Acc. & 0.045  & 0.046 & 0.051\\
Size &  101.26 & 111.86 & 112.46\\
\hline
\end{tabular}
\label{tab:num_adapter}
\end{subtable} \\
\begin{subtable}{0.49\linewidth}
\centering
\caption{LoRA's hidden dimension}
\begin{tabular}{c|ccc}
\hline
r & 2 & 4 & 8 \\
\hline
Acc. &0.051 &0.048 & 0.050 \\
Size &90.66  &90.74  & 90.88 \\
\hline
\end{tabular}
\label{tab:dimention_LoRA}
\end{subtable} 
\begin{subtable}{0.49\linewidth}
\centering
\caption{The number of LoRA}
\begin{tabular}{c|ccc}
\hline
& 1 & 2 & 3 \\
\hline
Acc. &0.048  & 0.049 & 0.047 \\
Size &90.74 & 90.88 & 91.08\\
\hline
\end{tabular}
\label{tab:num_LoRA}
\end{subtable}
\end{table}

\subsection{Generalization Performance}
To further demonstrate the superiority of the effectiveness and generalizability of the proposed SU-SAM, we compare it with the state-of-the-art approaches in Table~\ref{general}. From the table, we have two following observations:
Firstly, all the state-of-the-art approaches~\cite{singh2022cass,zhang2022dino,VST,zhuge2022salient,zhao2020suppress,DIS,jia2022segment,COD} show less-desirable performances in each dataset. Instead, our SU-SAM (specialized models) consistently outperforms these methods when fine-tuned on each downstream dataset (the best accuracy is shown in bold in  Table~\ref{general}). Secondly, almost all these state-of-the-art techniques are specifically designed for one task and cannot be generalized well to other tasks. Without any task-specific design, our proposed SU-SAM can be employed as a generalized model, which fine-tunes with all five downstream datasets in a unified, shared model. Importantly, unlike~\cite{VST,singh2022cass,zhang2022dino,codella2018skin}, we eliminate the need for employing additional techniques to further enhance the performance. As shown in Table~\ref{general}, our generalized model consistently promotes the performance of SAM on all datasets, showcasing its remarkable generalizability. \textit{More results can be seen in the supplementary material.}

\begin{table} [ht]
    \centering
       \caption{Comparison results (MAE) with state-of-the-art models on various segmentation tasks. }
       \vspace{1mm}
\resizebox{1.0\linewidth}{!}{
    \begin{tabular}{l|ccccc}
\hline
    Methods & ISIC~\cite{codella2018skin} & DUTS~\cite{DUTS} &  DIS-TE4~\cite{DIS} &   COD10K~\cite{COD}  & CHAME~\cite{skurowski2018animal}\\
 \hline
 \multicolumn{6}{c}{Specialized models}\\
 \hline
         CASS~\cite{singh2022cass} & 0.086 & -&  - &   -&     - \\
         DINO~\cite{zhang2022dino} & 0.081 & -& -  &   - &    -\\
        MSA~\cite{wu2023medical} & 0.049 & -& -  &   -&     - \\
         VST~\cite{VST} & -  & 0.037 & -  &  - &    -\\ 
         ICONet~\cite{zhuge2022salient}& -  &0.037 & -  &  - &     -\\
         Gate~\cite{zhao2020suppress}& -  & -&  0.109&  -&    - \\
         IS-Net~\cite{DIS}&  -  &-&  0.072&   - &    -\\
        SegMaR~\cite{jia2022segment} & -  & -  & -  & 0.034&     -\\
        SINet~\cite{COD} &-  & -  & -  & 0.092&   0.440\\
    \hline
         \multicolumn{6}{c}{the same framework, 4 specialized models}\\
    \hline
 Ours (SU-SAM) &  \textbf{0.044} & \textbf{0.029} & \textbf{0.049} & \textbf{0.025}   &   \textbf{0.033}\\
 \hline
 \multicolumn{6}{c}{generalized model}\\
 \hline
 SAM~\cite{SAM}& 0.419 & 0.298& 0.373 &0.217  &   0.334\\
 Ours (SU-SAM)  & 0.067  & 0.041 &0.067  & 0.067&   0.075\\
\hline
    \end{tabular} }
    \label{general}
\end{table}

\subsection{Efficiency Analysis of SU-SAM}

In practice, when selecting a particular model, the trade-off between accuracy and training cost is often considered rather than focusing solely on accuracy. Table \ref{param} presents a comprehensive analysis of the four variants proposed, taking into account the number of trainable parameters, their proportion to the total parameters of the original SAM model, and training efficiency. Parallel SU-SAM, Serial SU-SAM, and Mixed SU-SAM share the same adapter structure, with only differences in insertion positions, resulting in a similar increase in the model's parameter count but with slight variations in training efficiency. In comparison to the adapter method, LoRA variant has significantly fewer parameters, albeit with some accuracy loss. Because all the trainable parameters required by these methods are significantly fewer than the scale of the model itself, they exhibit comparable training speeds. Additionally, the parameter increase introduced by Mix SU-SAM, relative to the original SAM parameters (accounting for only 11.4\%), is deemed acceptable.

\begin{table}[t]
    \centering   
        \caption{Effect on the model size and training efficiency of the proposed variants on DUTS dataset.}
        \vspace{1mm}
\begin{tabular}{c|c|c| c}
    \hline
    Methods & Model Size (M)& Rate & Time (img/s)\\
    \hline
    Para &10.65M & 11.4\% & 2.64\\
    Series &10.65M & 11.4\% & 2.68\\
    Mixed &10.65M & 11.4\% & 2.58\\
    LoRA &0.15M & 0.16\% & 2.66\\
    \hline
    \end{tabular}
    \label{param}
  \end{table}
  \vspace{-2mm}

%% file: latex/5_conclusion.tex
\section{Conclusion}
\label{sec:conclusion}
In this paper, we propose a simple and unified framework called SU-SAM for adapting SAM in underperformed scenes. Our framework abstracts general modules of different methods into basic design elements, and we design four variants based on a shared theoretical framework. SU-SAM is research-friendly, efficient, modular, and scalable, which removes all dataset-specific designs and focuses solely on general optimization, ensuring that SU-SAM can be applied to all SAM-based and even Transformer-based models.
Experimental results on nine datasets have demonstrated that all four proposed variants significantly improve the inference performance of SAM on underperformed datasets. The Mix variant stands out with its remarkable performance on almost all datasets, demonstrating competitive results. In addition, although less frequently employed, the parallel architecture design has shown potential on certain datasets. 
Given the rapid advancement of vision foundational models and their extensive application to downstream tasks, our work provides valuable insights for future researchers to choose appropriate adapter insertion positions and methods based on practical needs.

%% file: latex/6_supple.tex
\section{Supplementary Material}

\subsection{More Implementation Details}
We refer to the introduced plugin as ``A'' (the entire code is provided in the supplementary).
We only insert ``A'' into the ``forward'' method of the ``Block'' function in the image encoder of SAM, once after the attention and once before the MLP. In the parallel structure, $X_{out} = X_{in} + A(X_{in})$, and in the 
 serial structure, $X_{out} = A(X_{in})$. 
Thus, we can seamlessly integrate SU-SAM into the original SAM architecture with minimal changes. 
During the model's forward pass, we only need to provide a few additional flags to choose between adapter and LoRA, and between parallel and serial structures.

\begin{figure*}[h]
\centering
\includegraphics[width=1.0\textwidth]{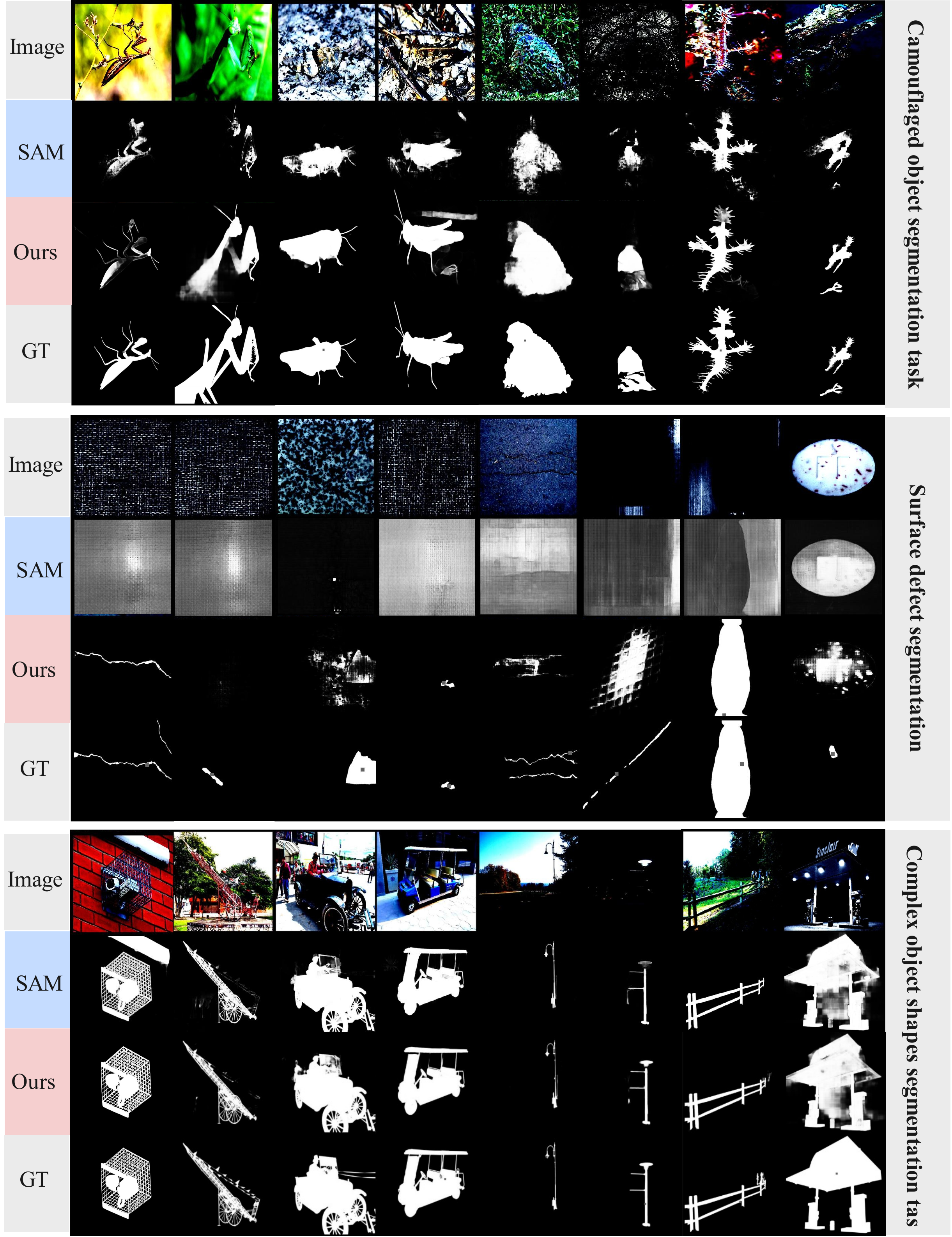} 
\caption{Visualization results on camouflaged object segmentation, surface defect segmentation, and complex object shapes segmentation.}
\label{DIS}
\end{figure*}

\begin{figure*}[t]
\centering
\includegraphics[width=1.0\textwidth]{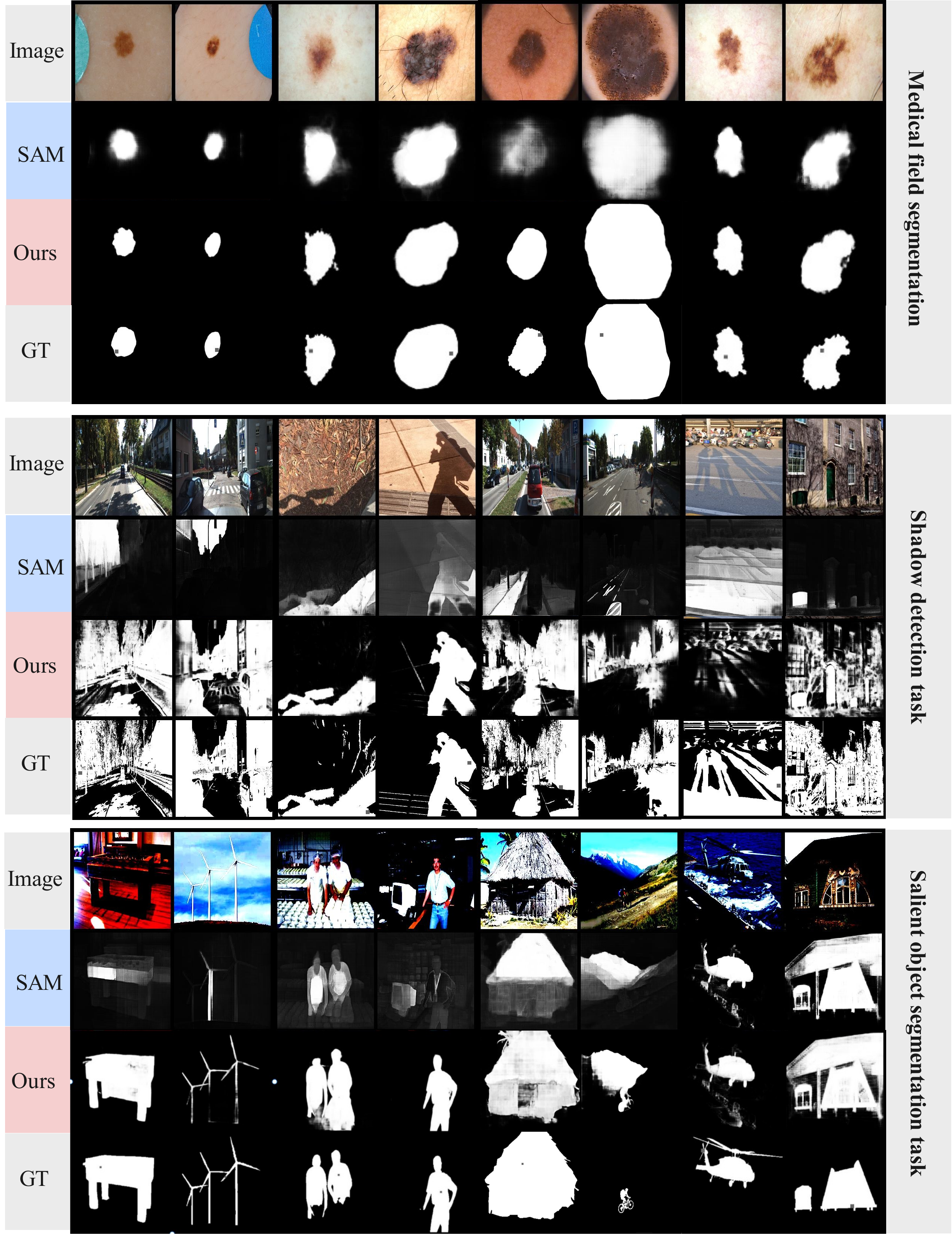} 
\caption{Visualization results on medical field segmentation, shadow detection, salient object segmentation. }
\label{COD10K}
\end{figure*}

To ensure a fair comparison, we maintain consistency in training hyperparameter settings unrelated to the methods. We follow work \cite{wu2023medical}, \cite{chen2023sam}, and \cite{zhang2023customized}. The modifications are solely made to the design dimensions of the main manuscript. We uniformly resized the images to 1024 without additional data preprocessing. The structure of all inserted adapters was fixed: a downsampling layer, a ReLU activation function, and an upsampling layer. The hidden dimension was set to 1/4 of the input dimension. To be more specific, the Adapter code is listed as follows: 
\begin{lstlisting}[language=Python, caption=Adapter Code, label=python-code]
class Adapter(nn.Module):
    def __init__(self, D_features, 
        mlp_ratio=0.25, 
        act_layer=nn.GELU, 
        skip_connect=True):
        super().__init__()
        self.skip_connect = skip_connect
        D_hidden_features =  
        int(D_features * mlp_ratio)
        self.act = act_layer()
        self.D_fc1 = 
        nn.Linear(D_features, D_hidden_features)
        self.D_fc2 = 
        nn.Linear(D_hidden_features, D_features)

    def forward(self, x):
        # x is (BT, HW+1, D)
        xs = self.D_fc1(x)
        xs = self.act(xs)
        xs = self.D_fc2(xs)
        if self.skip_connect:
            x = x + xs
        else:
            x = xs
        return x
\end{lstlisting}

We used the VIT-B SAM and Binary Cross-Entropy loss. AdamW optimizer was used for all experiments with an initial learning rate of 1e-4. The StepLR scheduler was employed with a decay step of 10 and a gamma value of 0.5. Fine-tuning was performed for 100 epochs on all datasets. All experiments were conducted on a single NVIDIA GTX 4090 GPU for fair comparisons.

\subsection{Visualization}
Following previous experimental protocol \cite{ji2023segment}, we verify our proposed method on the datasets where SAM is underperformed and demonstrate that our method could significantly improve the performance. The selected datasets encompass various downstream tasks, including salient object segmentation \cite{borji2019salient1}, complex object shapes segmentation \cite{DIS}, camouflaged object segmentation \cite{COD}, shadow detection \cite{shadow}, surface defect segmentation \cite{he2019fully}, and medical field segmentation \cite{tajbakhsh2015automated}. 
To showcase the efficacy of our approach, we present additional visual results as follows:

\subsection{Further Discussion}
SU-SAM is a well-encapsulated and straightforward design, making it easily adaptable and usable. For instance, it can be seamlessly integrated with optimization methods customized for specific datasets, including data pre-processing and post-processing. Additionally, it provides the flexibility to substitute SU-SAM's base adapter with a more intricate adapter design to enhance performance.  Considering that SU-SAM's base elements Adapter and LoRA have shown good performance in multimodal models, the design and discovery of SU-SAM can be used not only in large visual/multimedia models but also in other multimodal models. Importantly, due to the decoupled nature of our specific module design and the four insertion variants, additional modifications will not disrupt the normal operation of SU-SAM.